\documentclass[sigconf]{acmart}
\AtBeginDocument{%
  \providecommand\BibTeX{{%
    \normalfont B\kern-0.5em{\scshape i\kern-0.25em b}\kern-0.8em\TeX}}}

\RequirePackage[skins]{tcolorbox}
\newtcolorbox{custombox}[1]{
	colback=gray!10,
	colframe=gray!70,
	left=1mm,
	right=1mm,
	top=1mm,
	bottom=1mm,
	fonttitle=\bfseries,
	arc=0mm,
	leftrule=1mm,
	rightrule=0mm,
	toprule=0mm,
	bottomrule=0mm,
	notitle,
	before=\par\smallskip\noindent,
	before upper={\textbf{#1: } },
}
\usepackage{booktabs}

\setcopyright{rightsretained}
\copyrightyear{2023} 
\acmYear{2023}
\setcopyright{rightsretained}
\acmConference[ACE '24]{ACE '24}{January 29 - February 2, 2024}{Sydney, Australia} 
\begin{document}

\title[More Robots are Coming]{More Robots are Coming: Large Multimodal Models (ChatGPT) can Solve Visually Diverse Images of Parsons Problems}
%\title[More Robots are Coming]{More Robots are Coming: Multimodal Large Language Models (ChatGPT) can Solve Visually Diverse Images of Parsons Problems}

%\title[More Robots are Coming]{More Robots are Coming: Evaluating the Image Capabilities of GPT-4V and Bard in Solving Visually Diverse Parsons Problems}
%\title[More Robots are Coming]{More Robots are Coming: Comparing Bard and GPT-4V Image Capabilities to Visually Solve Parsons Problems}


% \author{Irene Y. Hou, Owen Man, Kenneth Angelikas, Sophie Metille, Stephen MacNeil}
% \affiliation{\institution{Temple University}
%             \department{Department of Computer and Information Sciences}}

% \author{Stephen MacNeil} % Steve 
% \email{smacneil@ucsd.edu}
% \affiliation{
%     \department{Department of Computer and Information Sciences}
%     \institution{Temple University}
%     \city{Philadelphia}
%     \state{Pennsylvania}
%     \country{USA}
% }

\author{Irene Y. Hou}
\email{irene.hou@temple.edu}
\affiliation{
  \institution{Temple University}
  \streetaddress{1801 N Broad St}
  \city{Philadelphia}
  \state{PA}
  \country{USA}
  \postcode{19122}
}

\author{Owen Man}
\email{owen.man@temple.edu} %\email{itsowen.dev@gmail.com}
\affiliation{
  \institution{Temple University}
  \streetaddress{1801 N Broad St}
  \city{Philadelphia}
  \state{PA}
  \country{USA}
  \postcode{19122}
}

\author{Sophia Mettille}
\email{sophia.mettille@temple.edu}
\affiliation{
  \institution{Temple University}
  \streetaddress{1801 N Broad St}
  \city{Philadelphia}
  \state{PA}
  \country{USA}
  \postcode{19122}
}

\author{Sebastian Gutierrez}
\email{guts@temple.edu}
\affiliation{
  \institution{Temple University}
  \streetaddress{1801 N Broad St}
  \city{Philadelphia}
  \state{PA}
  \country{USA}
  \postcode{19122}
}

\author{Kenneth Angelikas}
\email{kenneth.angelikas@temple.edu}
\affiliation{
  \institution{Temple University}
  \streetaddress{1801 N Broad St}
  \city{Philadelphia}
  \state{PA}
  \country{USA}
  \postcode{19122}
}

\author{Stephen MacNeil}
\email{stephen.macneil@temple.edu}
\affiliation{
  \institution{Temple University}
  \streetaddress{1801 N. Broad St}
  \city{Philadelphia}
  \state{PA}
  \country{USA}
  \postcode{19122}
}

\newcommand{\td}[1]{{\color{red} #1}}
\newcommand{\fb}[1]{{\color{blue} #1}}

\renewcommand{\shortauthors}{Hou, et al.}

\begin{abstract}

The advent of large language models is reshaping computing education. Recent research has demonstrated that these models can produce better explanations than students, answer multiple choice questions at or above the class average, and generate code that can pass automated tests in introductory courses. These capabilities have prompted instructors to rapidly adapt their courses and assessment methods to accommodate changes in learning objectives and the potential for academic integrity violations. While some scholars have advocated for the integration of visual problems as a safeguard against the capabilities of language models, new multimodal language models now have vision and language capabilities that may allow them to analyze and solve visual problems. In this paper, we evaluate the performance of two large multimodal models on visual assignments, with a specific focus on Parsons problems presented across diverse visual representations. Our results show that GPT-4V solved 96.7\% these visual problems, struggling minimally with a single Parsons problem. Conversely, Bard performed poorly by only solving 69.2\% of problems, struggling with common issues like hallucinations and refusals. These findings suggest that merely transitioning to visual programming problems might not be a panacea to issues of academic integrity in the generative AI era.

\end{abstract}


% Here is the table for all ACM CSS concepts https://dl.acm.org/ccs and I find Human-centered computing -> Visualization, Collaborative and social computing; Applied computing





%\keywords{problem framing, design repositories, design data, search facets} %TODO: needs iteration


%   \includegraphics[width=1\textwidth]{figures/teaserraw.png}
% \caption{Memory Sandbox is a system allowing end users 
%  to see and manage the memory of conversational agents. Memory Sandbox provides the following interaction affordances: 1) toggle memory visibility, 2) edit memory, 3) delete memory, 4) add memory, 5) summarize memory, 6) create conversation, and 7) share memory.}
%  % \Description{A website is presented with three main areas. The first is an area for writing problem statements. The second is an area for selecting stakeholders. The third is an area for viewing problem statements.}

\maketitle
\vspace{-0.2em}
\section{Introduction}

% Motivating the problem that LLMs introduce around assessment

Large language models (LLMs) are incredibly powerful tools with many exciting use cases emerging in computing education contexts. These models are now capable of generating code~\cite{chen2021evaluating, barke2022grounded, denny2023conversing, puryear2022github, wermelinger2023using}, answering multiple-choice questions~\cite{savelka2023generative, savelka2023thrilled}, and even solving Parsons problems~\cite{reeves2023evaluating}. While students can use these models to obtain high-quality explanations of code~\cite{macneil2022generating, macneil2023experiences, leinonen2023comparing}, the ability for students to use these tools to complete assignments or pass exams has raised numerous concerns from computing education researchers and practitioners about the effects these tools will have on assessment~\cite{macneil2023implications, prather2023transformed, prather2023robots, lau2023from, zastudil2023generative}.

%Computing education researchers have recently expressed both concern and excitement about the ways that generative models may affect the computing education landscape~\cite{macneil2023implications, macneil2022automatically, prather2023transformed, lau2023from, zastudil2023generative}. In all of these cases, assessment has been a central topic of discussion. 

% Work on parsons problems and visual assignments 

To address these concerns related to academic integrity, researchers have advocated for adopting proctored exams~\cite{zastudil2023generative, Joshi2023ChatGPT, rudolph2023chatgpt, prather2023robots, susnjak2022chatgpt} and visual programming problems~\cite{denny2023promptly} as potential barriers for students seeking these tools to cheat. However, drawbacks of in-person proctored exams and code interviews include scalability and known biases around test-taking~\cite{davis2013racial, holmes2021bad, xie2021domain}. Online AI-driven proctoring systems address the issues of scalability, while introducing additional concerns over technological accessibility in online exams, invasions of privacy and autonomy, as well as the risk of bias~\cite{gonzalez2020implementation, coghlan2021good, cahapay2021problems}. On the other hand, visual programming problems, such as `prompt problems'~\cite{denny2023promptly}, have been explored as another way to mitigate the use of generative AI; their design makes it challenging for computing students to directly feed into LLMs. Parsons problems are similarly challenging to input because existing user interfaces do not lend themselves to simple copy and pasting~\cite{reeves2023evaluating}. Instructors are even beginning to upload photos of their assignments with the hope that the effort associated with transcribing the text will be a barrier for students. While these approaches rely on visual reasoning to circumvent existing LLMs, the introduction of new large multimodal models, such as GPT-4V and Bard, are now capable of processing text and image data. 

In this paper, we investigate the performance of two large multimodal vision models (i.e.: GPT-4V~\footnote{https://cdn.openai.com/papers/GPTV\_System\_Card.pdf} and Bard~\footnote{https://bard.google.com/}) on solving Parsons problems across varying visual presentations as a first step in evaluating the capabilities of these new models. We created a dataset of Visual Parsons problems based on existing literature~\cite{reeves2023evaluating}, then converted each Parsons problem into diverse visual formats based on common Parsons Problem generation tools include JS-Parsons \cite{Sirkia2016ParsonVisulaization, helminen2012StudentsParson, Ihantola2011Parosn2D, Ihantola2013ProgrammingMobile}, Runestone~\cite{ericson2020Runestone, YeckehZaare2019SpacedTool, ericson2016IdentifyingDesign, ericson2015InteractiveEbook}, Epplets~\cite{kumar2018epplets, kumar2017effect}, and EvoParsons~\cite{bari2019evoparsons, gaspar2019lessons} . We then prompted GPT-4V and Bard with pre-engineered prompts to evaluate its ability in extracting Parsons problems from an image and solving them.

This paper investigates the following research questions: 
\begin{enumerate}
    \item [\textbf{RQ 1:}] How effectively do large multimodal models generate solutions to Parsons problems?
    \item [\textbf{RQ 2:}] How do visual variations of Parsons problems, recreated using different generation tools, influence the problem-solving performance of large multimodal models?
\end{enumerate}

Our results suggest that new large multimodal models (LMMs) are capable of solving Parsons problems across multiple visual representations with GPT-4V successfully solving 96.7\% of the problems. This paper contributes a critical new capability to the growing list of capabilities for LLMs, and now LMMs. This new capability is critical because it further threatens traditional assessment as students can use these models to now solve visual programming problems. However, a second LMM, Bard, was only able to solve 69.2\% of the problems. Furthermore, during our prompt engineering process, performance varied widely depending on the prompt. This reiterates the effect that the model and prompt have on performance.

\begin{figure}
    \centering
    \includegraphics[width=\linewidth]{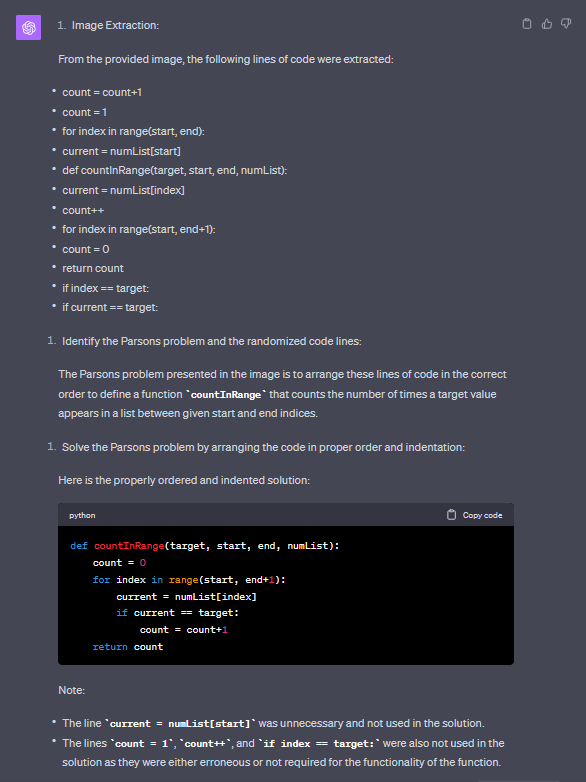}
    \caption{A screenshot of GPT-4V getting the correct answer for an image of a Parsons problem (EvoParsons Haynes-Magyar2022figure4). The response breaks down the problem solving process and even identified the distractors.}  
    \label{fig:gpt-answer}
\end{figure}

\section{Related Work} 

\subsection{Parsons Problems} 

Parsons problems are a type of programming puzzle where learners are given jumbled code snippets and must arrange them in the correct logical sequence rather than producing the code from scratch~\cite{parsons2006parson, ericson2022parsons, du2020review}. This approach has been explored and implemented in various educational settings, contributing to a deeper understanding of programming concepts~\cite{denny2008evaluating, ericson2017solving, ericson2022parsons, ericson2015InteractiveEbook, ericson2023conducting}. Parsons problems are often used to help novice learners learn early programming concepts, such as in CS1 and CS2 courses, by focusing on logic and program structure over the nuances of syntax~\cite{parsons2006parson}. 

Based on a recent ITiCSE working group focusing on Parsons problems, the most popular Parsons problem language was Python, and the most commonly targeted concepts in Parsons problems included loops, conditionals, and lists and arrays~\cite{ericson2022parsons}. Additionally, many variations of Parsons problems have been introduced to students: problems with distractors (code blocks not required for the solution)~\cite{denny2008evaluating}, `faded' Parsons problems (code blocks that include blanks)~\cite{harms2016distractors}, as well as ``two-dimensional'' problems, which require students to practice proper indentation~\cite{ihantola2011two}. Several Smart Learning Content providers (SLC)~\cite{brusilovsky2014increasing} have also created varied tools on their platforms for instructors to create and assess their own Parsons problems. These tools present Parsons problems in various visual formats, ranging from mobile apps~\cite{ihantola2013how} to online problem generators~\cite{ericson2019analysis, ihantola2013how, ericson2015InteractiveEbook}. 

A recent study examined the performance of LLMs at solving Parsons problems drawn from a review of the literature~\cite{reeves2023evaluating}. Their analysis, which focused on OpenAI Codex, showed that the model was only able to solve 51\% of the problems. Consequently, it is possible that Parsons problems may pose a challenge for LLMs. Moreover, Parsons problems are often difficult to copy and paste from the user interface into LLMs. Given these two challenges, it is possible that Parsons problems may discourage academic dishonesty given the low performance and need for transcribing the problems. Our research extends this prior work focusing on the effectiveness of the new large multimodal models (LMMs) which can analyze text and image data. %with LLMs due to their labor-intensive nature; however, the effectiveness of this deterrent is uncertain as visual multi-modal models enter the picture.

\subsection{The Effects of Generative AI on Assessment} 

Over the last two years, there has been an immense amount of research focusing on the capabilities of generative AI models. Initial research identified exciting new capabilities such as creating programming assignments for instructors~\cite{sarsa2022automatic, Finnie-Ansley2022Robots} and explaining code to students~\cite{macneil2022generating, macneil2023experiences, leinonen2023comparing}. As additional use cases have emerged showing how adept generative models are at solving programming problems~\cite{puryear2022github, wermelinger2023using}, automatically repairing bugs in code~\cite{koutcheme2023automated, fan2023automated, jiang2023impact}, and passing multiple choice quizzes~\cite{savelka2023can, savelka2023large, savelka2023thrilled}, researchers and practitioners have become concerned about the implications regarding assessment and academic integrity~\cite{zastudil2023generative, becker2023programming, lau2023from}. Initially, practitioners wanted to ban these models from their classrooms, but have since shifted toward the belief that `resistance is futile' — that students can and will use these models even if they are prohibited in the class~\cite{lau2023from}. An interview study with students corroborated these observations with multiple students saying they would use LLMs to help them with anything they considered `busy work'~\cite{zastudil2023generative}. This is also partially a reaction to the fact that there is currently no reliable way to differentiate between human and AI-generated content~\cite{sadasivan2023can}. %This is at least partly a reaction to recent papers that have demonstrated the difficulty in differentiating between human and AI-generated text~\cite{sadasivan2023can}.

\begin{figure*}[t]
    \centering
    \includegraphics[width=1\linewidth]{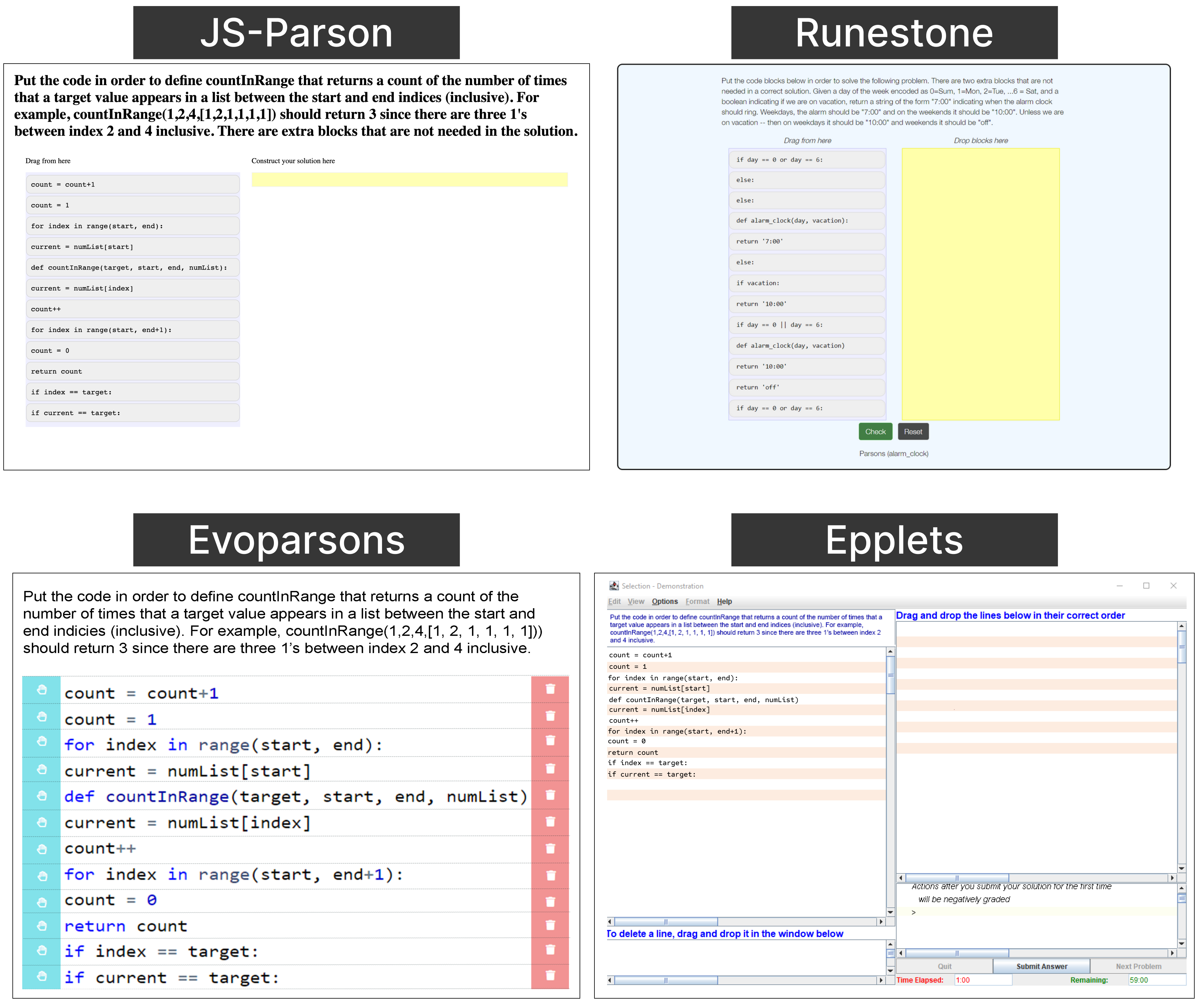}
    \caption{The four visual representations with the same coding example. The representations varied in aspect ratio, background color, fonts, font-size, text placement, and other visual nuances.}
    \label{fig:visual-problems}
\end{figure*}

To address these concerns around assessment, both researchers and practitioners have advocated for a return to heavily proctored exams~\cite{zastudil2023generative, Joshi2023ChatGPT, susnjak2022chatgpt}. While this may partially reduce the impacts of LLMs on assessment, students have shared that they strongly dislike this solution~\cite{zastudil2023generative} and proctored exams can disproportionately benefit students who are better test takers. Researchers have also devised `prompt problems' that make it harder for students to use generative AI tools directly~\cite{denny2023promptly}. However, with the introduction of large multimodal models, it is unclear whether this solution will continue to address the problem. Research is needed to understand the capabilities of LMMs given their potential impact on assessment.

\section{Methodology}

This study aims to investigate the recently deployed large multimodal models (LMMs) that generate natural language output given a text prompt and image. These models, such as Bard and GPT-4V, have not yet been studied in the context of computing education, but they could have significant impacts. Over a year ago, researchers raised the clarion call that `robots are coming'~\cite{Finnie-Ansley2022Robots}, but in this work we focus on the ability for these new models to interpret and solve visual programming problems. We use Parsons problems as a step toward understanding their capabilities and implications more broadly. To conduct this investigation, we needed to develop a dataset of visual Parsons problems and design a prompt that performs well at solving these problems. In this section, we describe how we developed our dataset, engineered an effective prompt, and formed the process for evaluating the model's performance.% on our dataset. 

\subsection{Visual Parsons Problems Dataset}

Given that a dataset of visual Parsons problems does not currently exist, we formed our own dataset guided by recent related research. Based on a recent working group by Ericson et al.~\cite{ericson2022parsons}, we chose four commonly referenced Parsons problem generators, which each varied in visual style. We then drew six Parsons problems from a recent study that assessed LLM performance on text-based Parsons problems~\cite{reeves2023evaluating}. The selected Parsons problems are representative of introductory programming concepts and have all appeared in previous literature.

\subsubsection{Four Visually Diverse Representations of Parsons Problems.}

Our dataset comprises four visually-diverse Parsons problem generation platforms, each offering distinct features and visual designs. Figure~\ref{fig:visual-problems} displays the visual differences between each Parsons problem generation tool, each option presenting varied colors, fonts, texts, formatting, user interface, and more. This selection of platforms was informed by a recent working group paper by Ericson et al.~\cite{ericson2022parsons}, which gave a comprehensive overview of leading Parsons problem platforms. After reviewing the most cited generation tools, we discarded the platforms that were not open-source or not readily available for public use. The finalized selected platforms included: JS-Parsons \cite{Sirkia2016ParsonVisulaization, helminen2012StudentsParson, Ihantola2011Parosn2D, Ihantola2013ProgrammingMobile}, Runestone~\cite{ericson2020Runestone, YeckehZaare2019SpacedTool, ericson2016IdentifyingDesign, ericson2015InteractiveEbook}, Epplets~\cite{kumar2018epplets, kumar2017effect}, and EvoParsons~\cite{bari2019evoparsons, gaspar2019lessons}. 

At a high level, all four platforms present one common goal to the user: arrange given code snippets in logical order according to instructions. However, the visual controls for achieving this goal vary. For instance, the JS-Parsons platform presents the problem at the top, with an interface below where the user must drag from a snippet bank on the left into a dedicated solution area on the right, with descriptions above each area. Runestone presents the problem similarly at the top with descriptions of each area, but also adds extra buttons for checking and resetting the work. EvoParsons also includes instructions at the top, yet introduces a unique single area of a list where the snippets all have symbolic icons that indicate dragging vertically to arrange or remove. Finally, Epplets presents a more complex visual landscape, including a full window UI with extraneous elements such as scroll bars, elapsed time, and menu options. This range of visual Parsons problem generators allows us to assess LMM performance in solving across various representations.

For each specified SLC platform, researchers transcribed each Parsons problem into the 4 visual representations. Each Parsons problem was transcribed identically, with scrambled ordering of code blocks remaining the same. Online platforms, such as JS-Parsons, allowed the researchers to input the Parsons problem line by line. For cases like Runestone, where problems are static, researchers utilized the web browser's Developer Tools to modify the HTML/CSS of the page, altering each code block individually to match the selected Parsons problem. Other platforms operated offline, so researchers used photo-manipulation tools to edit each problem's text and instructions, ensuring they mirrored the platform's appearance. Images of the Parsons problems were downloaded in the same PNG format to maintain uniformity in quality, compression, and other attributes.

\subsubsection{Six Parsons Problems}

To select the Parsons problems for our study, we aimed to include a set that represents the types of problems previously examined in computing education research. Remaining consistent with previous research, we relied on the six problems investigated by Reeves et al.~\cite{reeves2023evaluating} in a recent study of LLMs performance on Parsons problems. While previous work investigated these problems using text-based versions, we extend this work to investigate how large multimodal models perform on this task. 
The six Parsons problems from their study represent a variety of programming topics, including but not limited to loops, conditionals, and functions. To remain aligned with their study, we also kept all the problems in Python. This choice also corresponds with Python being the most commonly used language for representing Parsons problems~\cite{ericson2022parsons}. 
The problems varied not only in terms of their topic but also in their length and complexity, ranging from concise code snippets to more intricate programs. %This variety was intended to encompass real-world scenarios that students might encounter in solving Parsons problems. In using previous research as a benchmark we are able to test the same problems but in a new novel approach. 

\subsection{Model Selection}

Recently large multimodal models have been developed~\cite{wang2023review, liu2023hidden, gong2023multimodal, liu2023visual}. While new models are being released on a nearly monthly basis, we selected two highly popular models---OpenAI's GPT-4V and Google's Bard. %In our pilots, GPT-4V often outperformed Bard, which is consistent with recent findings~\cite{ali2023performace, bhardwaz2023Extensive}. 
At the time of writing this paper, GPT-4V is not yet accessible through the API. To ensure a fair comparison, we used the user interface to interact with GPT-4V and Bard.

\subsection{Prompt Engineering}

Prompt engineering is the iterative process of trying different inputs to an LLM in order to improve the performance. Our process started with the highly generic prompt ``solve the Parsons problem in the provided image.'' For this prompt, performance for GPT-4V was around 50\% and even lower for Bard. We issued multiple prompts for 10 random problem and representation pairs. From there, we iterated to improve a few key aspects: 1) reducing high refusal rates, 2) correctly extracting code blocks, and 3) getting the correct ordering and indentation. We were able to address these issues by explicitly explaining  in the prompt how Parsons problems work and that not all the code blocks were necessary. We also iterated on a few approaches including chain-of-thought prompting and few-shot prompting where examples were provided. These example-based techniques did not appear to improve performance in our testing. Our process involved developing prompts for both models, but we were only able to achieve high performance with GPT-4V. We selected the prompt that optimized performance for both models. 

The resulting prompt, shown below, consisted of three components which included the `Context` to inform the model that a screenshot of a coding Parsons problem was provided. The `Task` which described what a Parsons problem is and an overarching strategy for solving them. Finally, the `Instructions` explicitly listed the necessary steps in completing the task: 1) reading the text in the image, 2) identifying the task from the text, and 3) putting the snippets in the right order to complete the task. 

\begin{custombox}
 {Context}{I provided you with a screenshot of a coding Parsons problem.}
 \newline\newline
 \textbf{Task:} I want you to solve this Parsons problem. Each line of code is blocked and the order is randomized. Sort the lines of code in order and ensure unnecessary lines of code are not part of the answer.
 \newline\newline
 \textbf{Instructions:}
 \begin{enumerate}
    \item Image Extraction: Extract the text from the image
    \item Identify the Parsons problem and the randomized code lines
    \item Solve the Parsons problem by arranging the code in proper order and indentation
\end{enumerate}
\end{custombox}

We used the same prompt for both models. The goal was to ensure a consistent comparison of each model's responses across varied problems and representations. %By using an identical prompt for every trial, we aimed to ensure a consistent comparison of the model's responses across different Parsons problems. 

\subsection{Prompting Protocol} 

All responses were generated over a 3-day period starting on October 25, 2023. In total, there were 240 solution responses generated using GPT-4V and Bard. These 240 responses consisted of the 6 Parsons problems displayed across 4 visual representations with 5 attempts for each of the 2 models. %We issued multiple attempts to to account for the probabilistic nature of these models. 

\textbf{Generating Responses for GPT-4V and Bard:} We prompted each model with an image comprising the problem and representation pair accompanied by the final prompt shown previously. Each prompt and image was used as input five separate times to account for the probabilistic nature of these models. The chat window was closed and a new chat window was opened to avoid responses being affected by the chat history. The first response from the model was saved and analyzed. No follow-up prompting or clarifications were issued. So in cases where the model refused to respond, that was considered a wrong answer. This is important to note, because in our prompt engineering, we observed that follow up clarification prompts can reduce refusal rates and improve performance. When recording the results, three researchers assessed and discussed the validity of each response to ensure agreement. %We used the same input five time The order in which Parsons problems were inputted as new chat sessions into GPT-4V were consistent, and GPT-4V received no other priming outside the prepared prompt. In the event that GPT-4V made an error, the mistake was identified and no follow-up prompts were generated to rectify said error. Researchers proceeded only to prompt GPT-4V identically in a new chat session. When recording the results, three researchers assessed and discussed the validity of GPT-4V's answers to maintain inter-rater reliability. 

\textbf{Strict No-Regeneration Policy:} To minimize human variation, we adhered to a strict policy of never regenerating responses. Only the first response from the LMM was considered a solution and no follow-up prompting could be issued.

\subsection{Analysis} 

To evaluate the results, multiple members of the researcher team manually reviewed each output from the LMMs, with at least two members reviewing each output. A correct solution to a Parsons problem involves logically reordering the blocks with the proper indentation. Because each Parsons problem selected had a uniquely correct order, it was not necessary to execute the generated code. We did not observe any disagreements in evaluating the correct solutions between the two raters. 
In addition to this quantitative analysis, our team also qualitatively reviewed the outputs, looking for interesting outputs and trends. This open-coding method~\cite{strauss2004open} is common when there is a dearth of existing theories to guide the analysis deductively. The goal for this coding was to contextualize the findings with interesting behaviors from the model to shed light on nuances not captured in the quantitative analysis.

\section{Results}

%\subsection{Results} 

\begin{table*}
    
\centering
\begin{tabular}{@{}lccccc@{}}
\toprule
\textbf{Problem} &
  \multicolumn{4}{c}{\textbf{GPT4V: Number of Correctly Generated Solutions}} &
  \textbf{\% Total Success} \\ \midrule
\textbf{Image Source} &
  \multicolumn{1}{l}{Epplets} &
  \multicolumn{1}{l}{JS-Parsons} &
  \multicolumn{1}{l}{Runestone} &
  \multicolumn{1}{l|}{EvoParsons} &
  \multicolumn{1}{l}{} \\ \midrule
Weinmann2021figure1       & 5/5   & 5/5   & 5/5    & \multicolumn{1}{c|}{5/5}   & 100\%           \\
Haynes-Maygar2022figure2  & 5/5   & 5/5   & 3/5    & \multicolumn{1}{c|}{3/5}   & 80\%           \\
Ericson2022figure2        & 5/5   & 5/5   & 5/5    & \multicolumn{1}{c|}{5/5}   & 100\%           \\
Hou2022figure2            & 5/5   & 5/5   & 5/5    & \multicolumn{1}{c|}{5/5}   & 100\%           \\
Haynes-Maygar2022figure4  & 5/5   & 5/5   & 5/5    & \multicolumn{1}{c|}{5/5}   & 100\%           \\
Ericson2022figure4        & 5/5   & 5/5   & 5/5    & \multicolumn{1}{c|}{5/5}   & 100\%            \\ \midrule
\textbf{\% Total Success} & 100\% & 100\% & 93.3\% & \multicolumn{1}{c|}{93.3\%} & \textbf{96.7\%} \\ \bottomrule
\end{tabular}
    \vspace{1em}
    \caption{GPT-4V performance on 6 Parsons problems presented in 4 visual representations with the prompt issued 5 times for each problem and representation pair. Across the 120 responses, GPT-4V was correct 96.7\% of the time.}
    \label{tab:gpt-performance}
\end{table*}

\begin{table*}
    
\centering
\begin{tabular}{@{}lccccc@{}}
\toprule
\textbf{Problem} &
  \multicolumn{4}{c}{\textbf{Bard: Number of Correctly Generated Solutions}} &
  \textbf{\% Total Success} \\ \midrule
\textbf{Image Source} &
  \multicolumn{1}{l}{Epplets} &
  \multicolumn{1}{l}{JS-Parsons} &
  \multicolumn{1}{l}{Runestone} &
  \multicolumn{1}{l|}{EvoParsons} &
  \multicolumn{1}{l}{} \\ \midrule
Weinmann2021figure1       & 4/5   & 5/5   & 2/5    & \multicolumn{1}{c|}{5/5}   & 80\%           \\
Haynes-Maygar2022figure2  & 5/5   & 5/5   & 5/5    & \multicolumn{1}{c|}{5/5}   & 100\%           \\
Ericson2022figure2        & 0/5   & 5/5   & 4/5    & \multicolumn{1}{c|}{3/5}   & 60\%           \\
Hou2022figure2            & 5/5   & 5/5   & 5/5    & \multicolumn{1}{c|}{5/5}   & 100\%           \\
Haynes-Maygar2022figure4  & 0/5   & 5/5   & 4/5    & \multicolumn{1}{c|}{1/5}   & 50\%           \\
Ericson2022figure4        & 1/5   & 0/5   & 0/5    & \multicolumn{1}{c|}{4/5}   & 25\%            \\ \midrule
\textbf{\% Total Success} & 50\% & 83.3\% & 66.7\% & \multicolumn{1}{c|}{76.7\%} & \textbf{69.2\%} \\ \bottomrule
\end{tabular}
    \vspace{1em}
    \caption{Bard performance on 6 Parsons problems presented in 4 visual representations with the prompt issued 5 times for each problem and representation pair. Across the 120 responses, Bard was correct 69.2\% of the time.}
    \label{tab:bard-performance}
\end{table*}

We evaluated the performance of GPT-4V and Bard with 6 Parsons problems across 4 visual representations and with 5 attempts for each problem-representation pair. This resulted in a total of 120 responses per model with 240 total responses. Based on a McNemar test, which accounted for paired responses, there was a significant performance difference between Bard and GPT-4V conditions ($\chi^2$ = 24.976, df = 1, p < 0.001). The percentage of questions correctly answered for each model was 69.2\% for Bard and 96.7\% for GPT-4V.

\subsection{RQ 1: Correctness of Generated Solutions}

\subsubsection{Correctness of GPT-4V Generated Solutions}

\begin{figure}
    \centering
\includegraphics[width=0.95\linewidth]{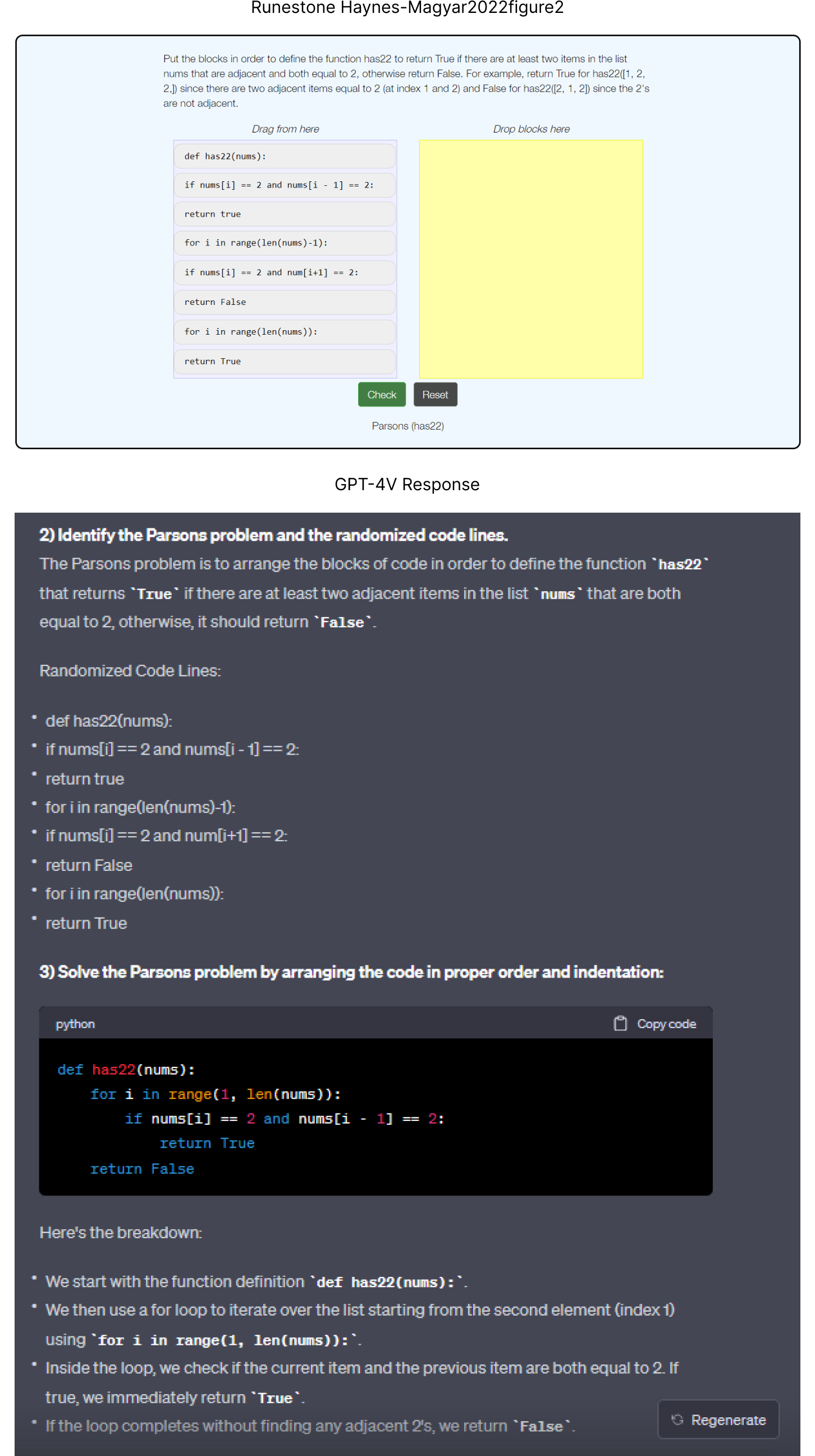}
    \caption{Runestone visual representation of the Haynes-Magyar2022figure2 Parsons problem. In its response, GPT-4V incorrectly alters a code block to "for i in range(1, len(nums))."}
    \label{fig:fullericson}
\end{figure}

Table~\ref{tab:gpt-performance} summarizes the performance of GPT-4V across 5 attempts for each of the 6 Parsons problems displayed in all 4 visual representations. GPT-4V successfully solved 116 out the of the 120 attempts, achieving an overall success rate of 96.7\%. The only mistakes that GPT-4V made were on the Haynes-Maygar2022figure2 problem. Shown in Figure~\ref{fig:fullericson}, GPT-4V hallucinated and rewrote one of the code blocks.% as shown in Figure~\ref{fig:fullericson}.

GPT-4V appeared to correctly extract and identify the code blocks with the exception of the Haynes-Maygar2022figure2 problem. In that case, it appears highly likely that it extracted the code block correctly, given that the hallucination is semantically equivalent. 
%GPT-4V was able to correctly extract and identify the code blocks in 100\% of the cases. 
When distractor code blocks were included, it always explicitly identified them, as shown in Figure~\ref{fig:gpt-answer}. Whether or not the explicit identification of distractors played a role in the overall performance, the implication is that the model can identify and ignore plausible distractors when included. It is also worth noting that there were syntactic variations in how GPT-4V produced conversational text for each response, although the final code solution remained consistently correct. In some cases, it explained the code blocks, in other cases it explained the reasoning such as what is needed to solve the problem. It seldom generated test cases. %In a number of cases it presented the lines of code in a code block with the programming language mislabeled. }

\subsubsection{Correctness of Bard Generated Solutions}

Bard only produced a correct solution on 69.2\% of the attempts. This lower performance appeared to be influenced by a number of factors including refusing to answer (7.5\% refusal rate), hallucinating and editing the code blocks (9.6\%), producing incorrect sequencing (3.3\%), or other problems. Aligned with prior work, we observed cases where Bard struggled with the concept of negation~\cite{tran2023generating}.

In terms of hallucinations, Bard had a tendency to edit code blocks. In some cases, Bard converted between single quotes and double quotes or added spacing that was not in the original code block. The correctness of these solutions was assessed on a case by case basis as it often did not affect the correctness of code. In more extreme cases, the code block was rewritten. For example, in the EvoParsons Haynes-Maygar2022figure4 problem, Bard rewrote a code block 4 out of 5 times to be ``count +=1'' rather than ``count = count + 1.'' Similarly to GPT-4V, there were also variations in the structure and presentation of the solution. In some cases, Bard explained the produced code and in other cases it generated test cases. It occasionally solved the problem by first explicitly identifying the code blocks and then by composing them into a solution. 

Finally, Bard performed better on some problems than others. The multiple regression analysis which modeled both the problem and the representation as independent variables revealed a significant overall effect (F-statistic = 10.02, $p < 0.001$) with an adjusted $R^2$ of 0.3776, indicating that the predictors collectively explain a substantial portion of the variance in scores. After correcting for multiple comparisons, the coefficients for individual predictors suggested that EricsonFig4 (Estimate = -0.35, $p < 0.01$), HaynesFig2 (Estimate = 0.40, $p < 0.001$), and ProblemHou2022 (Estimate = 0.40, $p < 0.001$) had significant effects on scores compared with the baseline of Ericson2022figure2. While various factors may contribute to these performance differences, the findings indicate that performance indeed varies across different problem types.

\subsection{RQ 2: Effects of Visual Representation}

\subsubsection{Effects of Visual Representation on GPT-4V Performance}

GPT-4V appeared to recognize the text correctly without refusals (e.g. ``I can not seem to extract the text from the image''). The only issues we observed were cases where GPT-4V hallucinated code that was nearly semantically identical as shown in Figure~\ref{fig:fullericson}. This suggests that the code was recognized and understood but there may have been issues in terms of how it was encoded in the vector space or how the generated text was produced.  

\subsubsection{Effects of Visual Representation on Bard Performance}

\begin{figure}
    \centering
\includegraphics[width=0.95\linewidth]{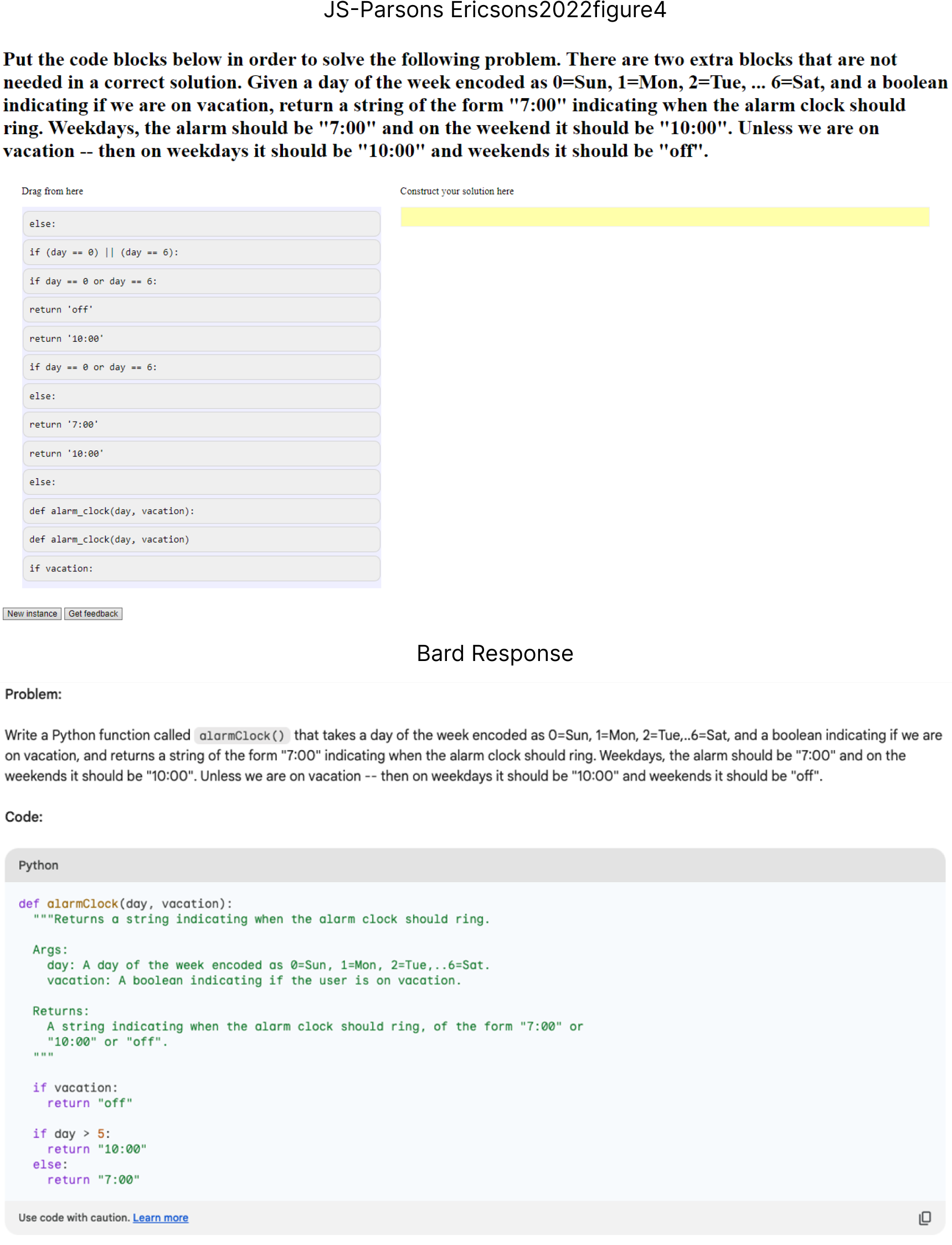}
    \caption{JS-Parsons visual representation of the Ericson2022figure4 Parsons problem. In its response, Bard extracts the text incorrectly and generates its own code blocks.}
    \label{fig:incorrect-ericson-bard}
\end{figure}

Bard struggled significantly more than GPT-4V in image-to-text extraction. In 7.5\% of all trials, Bard refused to solve the problem. For the Ericson2022figure4 problem in the Epplets format, Bard refused to attempt to solve the problem in all 5 trials. For those cases, the response included a variation of the phrase  ``I'm only a language model'', which could mislead users to believe that it does not have capability to process image data. %in any capacity, stating, ``I'm only a language model and don't have the capacity to understand and respond.'' 
Based on the multiple regression analysis described earlier, which accounted for multiple comparisons, visual representation had an impact on the success rate, with the JS Parsons representation showing significantly higher scores than Epplets (Estimate=0.33, $p<0.01$). In the worst case, Bard was only able to solve 50\% of the problems when represented in the Epplets visual format. In cases such as in Figure \ref{fig:incorrect-ericson-bard}, Bard failed to solve the problem because it hallucinates incorrect code blocks, rewriting lines such as ``if day == 0 or day == 6'' into ``if day > 5.'' Given these are both conditional statements, this suggests a semantic understanding issue rather than a problem with text recognition.

\section{Discussion}

\subsection{RQ1: Correctness of Generated Solutions}

The results of our study suggest that large multimodal models (LMMs) are already capable of solving visual programming problems with remarkable proficiency. We investigated this capability with Parsons problems presented in four visually diverse representations. There are two main aspects of solving visual programming problems: (1) analyzing the image and (2) solving the coding problem. GPT-4V performed well across the range of visual presentation formats, despite variations in fonts, font sizes, backgrounds, colors, interaction styles, and text location. 

GPT-4V was able to solve 96.7\% of all Parsons problems on its second attempt, providing consistently perfect indentation and syntax. Bard was only able to solve 69.2\% problems, which was still substantially higher than the 51\% correctness achieved in previous studies when OpenAI Codex was used~\cite{reeves2023evaluating}. This is noteworthy, because unlike in that prior work, all of the Parsons problems that we evaluated were presented to the model without the help of pre-indentation. In addition to using a more recent model, our prompt was also more explicit, with included context, task, and step-by-step instructions. During prompt engineering, the performance was closer to 50\% if we simply asked GPT-4V to ``solve the Parsons problem in the provided image.'' Another observation is that the type of problem, including problems that contained distractors, did not appear to negatively impact the performance of GPT-4V. Bard encountered difficulties with specific problems.

\subsection{RQ2: Effects of Visual Representation}

% DRAFT
The Parsons problems were presented in multiple diverse visual representations. While this did not appear to affect GPT-4V, we observed that some visual representations were easier for Bard than others. For Bard, problems presented in Epplets format had the lowest performance. While it is unclear what about Epplets led to the lower performance, visual inspection suggests that it has the smallest font size and more visual stimuli than the others. The interface also contained more text input windows than the other representations. There was a window containing the whole program, the fragment bank, the solution area, and two other irrelevant windows at the bottom containing additional text. To add to the complexity, each window had both horizontal and vertical scroll bars. There were buttons at the bottom right of the image indicate actions for quitting, submitting, and navigating to a different problem. A menu bar at the top left included various options, editing, viewing, formatting, and help drop-downs. In addition to the input controls, instructions related to the program itself were visible in the interface. There were two timers in the bottom right of the input that indicate the elapsed time and remaining time. Finally, a message related to grading appears in the bottom right window. GPT-4V did not appear to be affected by the visual representation, but these visual elements negatively affect the performance of Bard.

\subsection{Implications for Assessment}

Recently, researchers and practitioners have been investigating techniques to address unsanctioned use of large language models in computing classrooms, such as presenting problems visually~\cite{denny2023promptly}. However, our results suggest that LMMs are now capable of solving programming problems that are represented visually. This introduces a new threat toward traditional assessment and challenges educators to reconsider assessment in the age of generative AI. 

There appear to be two paths forward. The first is adopting adversarial methods toward assessment which could include ``data poisoning'' attacks~\cite{shan2023prompt}. Data poisoning or `AI poisoning' is a technique to ``manipulate training data to introduce unexpected behaviors into machine learning models at training time.'' These attacks focus on training data, but they could also focus on input. Our results show some preliminary evidence for this approach. Since Bard faced challenges with Epplets due to visual complexity, creating visually intricate problems with distracting elements might deceive models to perform poorly. However, this may also confuse students and negatively impact learning. A variation could be to inject noise that is not perceptible to students, but that would cause the model to perform poorly. This is likely a temporary solution given the ongoing improvements in model performance.

A second path is to consider alternative forms of assessment and relying on intrinsic motivation to ensure students do not use models in unsanctioned ways. This is an exciting approach because ungrading~\cite{blum2020ungrading, chu2020, spurlock2023improving}, culturally responsive computing~\cite{scott2015culturally}, and other alternative pedagogical styles~\cite{latulipe2015structuring, macneil2016exploring} have shown great promise, but have not yet been widely adopted. Perhaps these new threats to assessment may challenge educators to try them in their classrooms.

%Finally, there are also some intriguing findings that add some nuance to this story. 

\subsection{Implications for Pedagogy} 

This paper's results add to a growing list of examples showcasing how LLMs and LMMs can effectively solve problems for students~\cite{macneil2022generating, macneil2023experiences, leinonen2023comparing, puryear2022github, wermelinger2023using, koutcheme2023automated, fan2023automated, jiang2023impact, savelka2023can, savelka2023large, savelka2023thrilled}. While there are clear implications for assessment, there are also considerations for pedagogy, especially around the ethical use of generative AI. 

Our results show that some models perform better than others. Based on our prompt engineering, we also reiterate previous work about the importance of identifying effective prompts. Taken together, these observations suggest that some students may be able to use these models to great effect, while others may struggle to use them to get the correct answer. This suggests that there is a potential for unequal benefits. Additional work is needed to understand student prompting behaviors~\cite{zamfirescu2023johnny}.

\section{Limitations} 

 %This led to a smaller dataset compared to previous work~\cite{reeves2023evaluating}. Future work might expand to include a broader spectrum of Parsons problem difficulty and a more extensive set of visual representations.

There are two limitations that must be considered when evaluating the results presented in this paper. First is the number of visual representations presented and second is the selection of Parsons problems rather than other visual programming problems. 

% Visual representation

The first limitation was related to the availability and accessibility of diverse visual representations of Parsons problems. This constraint stemmed from several factors, including the scarcity of open-source Parson problem generators and the lack of existing visual Parson problem datasets. Another limitation is our approach to obtaining visual representations, which relied on capturing screenshots and downloading Parson problems. This led to some minor variations in PNG aspect ratios. Despite efforts to standardize, such as downloading images in the same file formats and with the same pixel density, these minor discrepancies remained. While this reflects a good sampling of the real-world diversity of visual representations students might encounter, it may introduce the possibility that GPT-4V's performance varies depending on additional factors such as aspect ratio or pixel density. 

% Parsons problems vs other problems 
The second limitation is our focus solely on Parson problems as a form of visual programming problems. While these problems have provided valuable insights, it is essential to recognize that there are many other visual programming problems, such as UML diagrams, input-output tables, and data-structures problems involving graphs and trees. %,  such as concept maps~\cite{keppens2008concept}, interactive simulations~\cite{vogel2006computer}, and virtual labs~\cite{lynch2017review}, which also play crucial roles in education. 
As future work, we are evaluating these alternative problems to provide a more comprehensive understanding of the performance of these models on diverse visual programming problems. Furthermore, we acknowledge that our study exclusively relied on only two LMMs, GPT-4V and Bard. While these models are extremely popular, it is essential to recognize that other LMMs are available~\cite{wang2023review, gong2023multimodal, liu2023visual}. Assessing the performance of LMMs would have offered a broader perspective on their respective strengths and weaknesses. Such an approach could have illuminated how these models perform in various scenarios and their overall efficacy in the broader educational landscape.

\section{Conclusion}

In conclusion, the rise of LLMs has prompted educators to reassess their assessment methods and devise new ways of ensuring students `do the work.' To address this issue, researchers and practitioners have proposed assigning visual programming problems to make it harder for students to use LLMs. Our study examined the performance of the recently released large multimodal models (LMMs) on multiple Parsons problems presented across diverse visual representations. GPT-4V and Bard were able to solve 96.7\% and 69.2\% respectively. The remarkable proficiency demonstrated by GPT-4V in tackling these problems across differences in visual presentation, challenges the idea that using visual programming problems will be a panacea for maintaining traditional assessment in the era of Generative AI. As we navigate this evolving landscape, it becomes clear that broader pedagogical considerations are essential to effectively address the impact of these models and ensure a holistic and adaptive approach to teaching and learning. First, we recommend that instructors reconsider their assessment practices in light of the growing number of AI capabilities, and second, we recommend that instructors consider the wide range of performance obtained by LMMs based on how they are prompted and which model is selected. These provide two interesting directions for future work.

\balance
\bibliographystyle{ACM-Reference-Format}
\bibliography{sample}


\end{document}